\documentclass[journal]{IEEEtran}
\usepackage{amsmath,amssymb}
\usepackage{psfrag}
\usepackage{epsfig}
\usepackage{cite}
\usepackage{graphics}
\usepackage{color}
\usepackage{subfigure}
\usepackage{multirow}
\usepackage{threeparttable}
\usepackage{booktabs}
\usepackage{mathrsfs}
\usepackage{bm}
\usepackage[ruled,vlined]{algorithm2e}

\usepackage{hyperref}

\begin{document}

\title{BSAS:Beetle Swarm Antennae Search Algorithm for Optimization
Problems}

\author{Jiangyu Wang, Huanxin Chen*}


\markboth{}%
{Shell \MakeLowercase{\textit{et al.}}: Bare Demo of IEEEtran.cls
	for Journals} \maketitle
\begin{abstract}
Beetle antennae search (BAS) is an efficient meta-heuristic algorithm.
However, the convergent results of BAS rely heavily on the random beetle
direction in every iterations. More specifically, different random seeds
may cause different optimized results. Besides, the step-size update
algorithm of BAS cannot guarantee objective become smaller in iterative
process. In order to solve these problems, this paper proposes Beetle
Swarm Antennae Search Algorithm (BSAS) which combines \emph{swarm
intelligence algorithm} with \emph{feedback-based step-size update
strategy}. BSAS employs \emph{k} beetles to find more optimal position
in each moving rather than one beetle. The step-size updates only when
\emph{k} beetles return without better choices. Experiments are carried
out on building system identification. The results reveal the efficacy
of the BSAS algorithm to avoid influence of random direction of Beetle.
In addition, the estimation errors decrease as the beetles number goes
up.
\end{abstract}

\section{Introduction}\label{introduction}
Meta-heuristics have enormous potential for solving complex optimization problems where analytical solutions are not available.
Jiang and Li\cite{ref.BAS} develop a algorithm called Beetle antennae search(BAS), which is taking inspiration from detecting and searching behavior of longhorn beetles. BAS shows good performance in the benchmark experiments which aims at searching global optimum value of two typical test functions, namely Michalwwicz and Goldstein-Price.

However, the results deviate greatly while we try to identify a high-dimensional system, for example, a lumped resistance-capacitance model with nine unknown parameters. There are two reasons why BAS performs not well when dealing with this optimization problem.

On the one hand, the convergence results of  BAS in \cite{ref.BAS} is highly depentdent on the beetle direction(random walking mechanism in \cite{ref.BAS}), which is randomly generated in each iteration, thus causing unstable results when faced with complex and high diamensional problems. On the other hand, the step size of each iteration will be attenuated. In other words, the step size is updated  regardless of the objective function value is better or not. In general, due to the random direction and the inefficient step size update strategy, the BAS may converge early and fall into the local optimal solution for high-dimensional problems. What's more, the results of multiple tests on the same objective function may vary greatly.

Aim at solving aforementioned problems, this paper proposes a modified BAS algorithm, namely Beetle swarm antennae search(BSAS), which takes \emph{swarm intelligence} and \emph{feedback-based step update strategy} into consideration.  In this paper, BSAS is applied to estimate parameters (including the initial temperature value) for a resistance-capacitance(RC) model, which is widely used in describing the thermal dynamics of the buildings . The main contributions of this paper are illuminated as below:
\begin{itemize}
	\item Feedback-based step update strategy is applied to make sense of every move of beetle. In other word,  the step-size updates only if the next hypothetical position of beetle cannot make the objective function value better.
	\item Beetles swarm is employed to search possible positions which optimizes the objective function in each iteration, thus overcoming the blindness of random walking of one beetle in a certain degree.
\end{itemize}

The rest of the paper is organized as follows. In Section \ref{Prposed -approach-design}, BSAS is proposed. In Section \ref{validation}, BSAS is applied to system identification of a RC model and the numerical results are presented and compared to the results of BAS. Section \ref{conclusion} concludes the main works of this paper.

\section{Proposed approach}\label{Prposed -approach-design}
In this section, an improved BAS is proposed by  taking both of swarm intelligence algorithm with feedback-based step-size update
strategy into consideration. Besides,  two diagrams describing optimization process of BAS and BSAS respectively are given to illuminate the difference of two algorithms and make them easily understood.

\subsection{Original BAS}\label{the-original-BAS}
The original BAS includes two rules, namely seraching behavior and detecting behavior. In each timestep, beetle moves in a random direction. Therefore, a random unit vector $\overrightarrow{\bm{b}} \in \mathbb{R}^N$ is used to describe the direction. The 
coordinate of  both right-hand and left-hand sides of the beetle's antennae are presented as (\ref{xrxl}),
\begin{eqnarray}\label{xrxl}
\bm{x}_r&=&\bm{x}^t+d^t\overrightarrow{\bm{b}}, \nonumber\\
\bm{x}_l&=&\bm{x}^t-d^t\overrightarrow{\bm{b}},
\end{eqnarray}
where the subscript $r,l$ represents the right-hand side and left-hand side respectively, the superscript $t$ represent specific moment. $d$ is behalf of  distance from the antenna to the centroid of the beetle. More specfically, $d^t$ represents corresponding beetle's searching area at a certain moment, and the area will attenuate over time as described in (\ref{d.update}). $d_0$ guarantees that $d^t$ will not reduce to zero. By the way, magnitude of $d_0$ depends on the scale of specific problem. The beetle's detecting behavior is described as (\ref{x.update}),
\begin{equation}\label{x.update}
\bm{x}^t=\bm{x}^{t-1}+\delta^t\overrightarrow{\bm{b}}\text{sign}(f(\bm{x}_r)-f(\bm{x}_l)),
\end{equation}
where $\delta^t$ represents the step size, which will reduce over time as demonstrated in (\ref{delta.update}).
 The update rules of $d$ and $\delta$ are presented in (\ref{d.update}) and (\ref{delta.update}). 
\begin{eqnarray}
d^t&=&\eta_{d}d^{t-1}+d_0,\label{d.update}\\
\delta^t&=&\eta_{\delta}\delta^{t-1} + \delta_0.\label{delta.update}
\end{eqnarray}
where $\eta_{d}$ and $\eta_{\delta}$ are attenuation coefficients of antennae length $d$ and step size $\delta$ respectively.
 Jiang and Li \cite{ref.BAS-WPT} indicate that presetting of parameters such as $d$ and $\delta$ influences performance of BAS seriously. In Section\ref{validation}, we discuss about parameters initialization in brief for the concerned problem,namely RC model estimation, in this paper. 

\begin{figure}[ht]
	\centering 
	\includegraphics[width=0.9\linewidth]{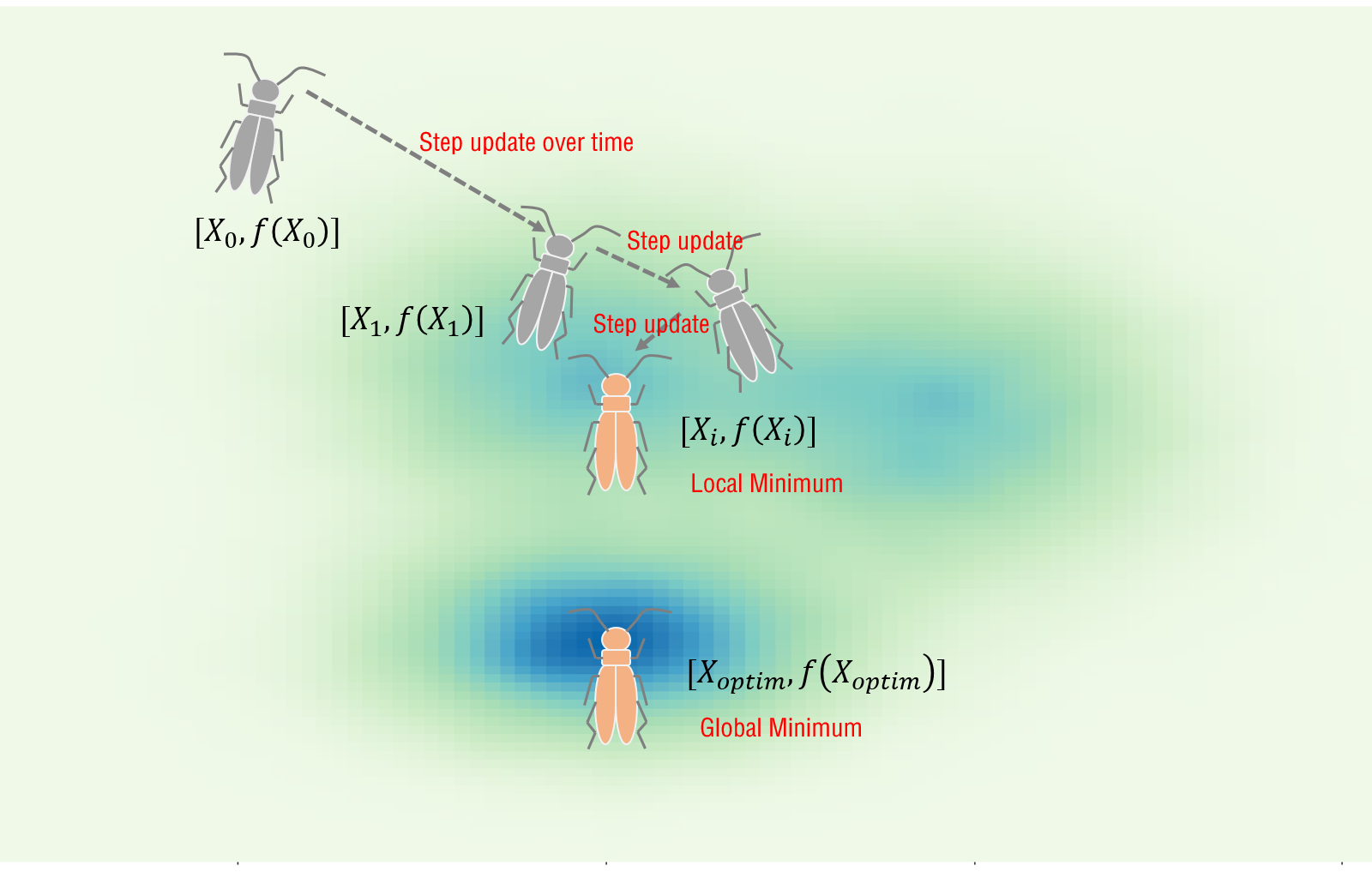}  
	\caption{Step updating in BAS process. The convergence results is not global optimum because the step decrease over time instead of objective feedback..}  
	\label{fig:BAS}  
\end{figure}

The diagrammatic drawing of BAS algorithm is shown in Fig.\ref{fig:BAS}. Beetle moves according to a random direction in every iteration. Meanwhile, the position updates as (\ref{x.update}). The grey dotted line with arrow  points out the direction. Evidently, the length of grey line, which represents the step-size, reduces over time as (\ref{d.update}). Note that Fig.\ref{fig:BAS} doesn't show the process of $\delta$ updating to make the beetle big enough to observe.

We can see that only one position in the random direction is explored in one iteration. In addition,  beetle position and the step size updates regardless of the objective function value obtained in last iteration is smaller or larger. A possible scenario is shown in Fig.\ref{fig:BAS}, in which the step size reduces too fast and the directions seem not good in some iterations. As a result of this scene, the beetle fall into local optimum position shown in wathet blue and cannot jump out of the local area. 

\subsection{Proposed BSAS algorithms}\label{the-BSAS}

In general, there are two drawbacks for BAS algorithm. First,  beetle moves in onerandom direction for each iteration. It cannot guarantee that the move of beetle will  make the objective function value better. Correspondingly,  we can improve the searching behavior of BAS algotithm by employing $k$ beetles to move in $k$ directions, thus enhancing the possibility for finding better position of beetle. Second, every movement of the beetle will cause an update of the position and step size. In fact, if the objective function is better, the position of beetle should be updated while the step size not. Vice versa. Therefore, we can design a feedback-based position and step-size update strategy as shown in algorithm \ref{BSAS.optim}.

\begin{figure}[ht]
	\centering  
	\includegraphics[width=0.9\linewidth]{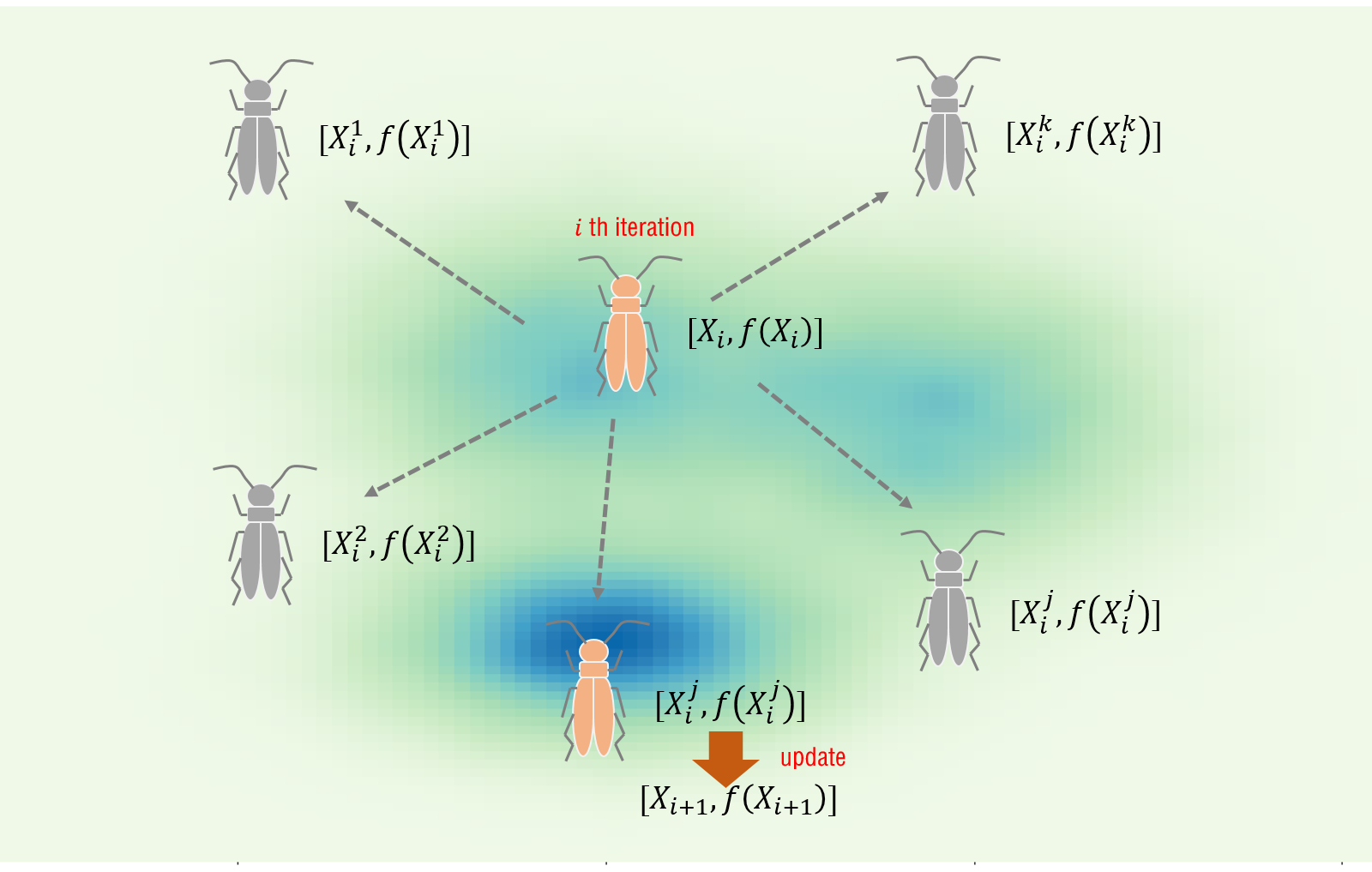}  
	\caption{Position updating process in BSAS. In every iteration, $k$ beetles are employed to find better parameters set. The beetle position updates until one of the beetles find better parameters set to optimize the objective function.}  
	\label{fig:BSAS}  
\end{figure}

In addition, a probability constant $p_{\delta}$ is introduced to measure the impact of random directions. In other word, we think there is a small probability $p_{\delta}$ that $k$ beetles will miss the better position (parameters set) for objective function at the current step size. Therefore, we generate a random number which belongs to $[0,1]$ and compare it to $p_{\delta}$. As a result, in most of the time, if $k$ beetles cannot find smaller objective function value, BSAS algorithm thinks there are no better positions at current step size and it needs to be updated. In a few cases, the algorithm thinks there are still some positions to satisify beetle's need to minimize the objective, but these beetles cannot find it because of limited beetles number $k$. Therefore, step-size $\delta$ and sensing lenght $d$ will remain the same if the random number is less than $p_{\delta}$. However, the beetle position updates only if bettles swarm find better positions to optimize objective function as shown in Fig.\ref{fig:BSAS}.
\begin{algorithm}\label{BSAS.optim}
	\caption{BSAS algorithm for global minimum searching}
	
	\KwIn{Establish an objective function $f(\bm{x}^t)$, where variable $\bm{x}^t=[x_1,\cdots,x_n]^\text{T}$ , initialize the parameters $\bm{x}^0,d^0,\delta^0,k,p_{\delta}$.}
	\KwOut{$\bm{x}_\text{bst}$, $f_\text{bst}$.}
	
	\While{ ($t<T_\text{max}$) or ($\delta < \delta_\text{criterion}$)}{
		Generate $k$ direction vector units $[\overrightarrow{\bm{b}_1} , \overrightarrow{\bm{b}_2}, \cdots,\overrightarrow{\bm{b}_k}]$ \;
		Search in variable space with two kinds of antennae according to (\ref{xrxl}) for $k$ beetles \;
		\eIf{$min(f(\bm{x}_i^t))<f_\text{bst},\text{where } i \in [1,2,\cdots,k]$}
		{$f_\text{bst}=f(\bm{x}^t)$\;$\bm{x}_\text{bst}=argmin(f(\bm{x}_i^t))$\;$\bm{x}^{t+1}=\bm{x}_\text{bst}$.}
		{\eIf{$rnd(1) > p_{\delta}$}
			{Update sensing diameter $d$ and step size $\delta$ with decreasing functions (\ref{d.update}) and (\ref{delta.update}) respectively\;}
			{Parameters $\delta$ and $d$ remain the same.}}
	}
	
	\Return{$\bm{x}_\text{bst}$, $f_\text{bst}$.}
\end{algorithm}

\section{Validation}\label{validation}
In order to verify the efficacy of the proposed BSAS algorithm for high-dimensional and complex system, RC model is applied to validate the improved algorithm. In this section, we use both of BAS and BSAS algorithm to estimate the parameters of RC model. Then the numerical results are compared.
\subsection{A. Test model}\label{test-model}
\begin{figure}[ht]
	\centering  
	\includegraphics[width=0.9\linewidth]{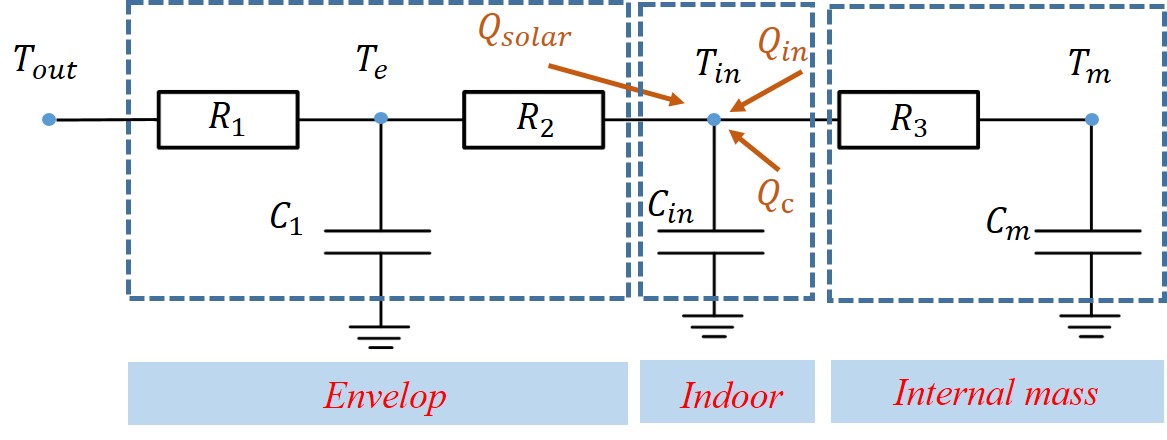}  
	\caption{The lumped resistance-capacitance model of a one-zone building.}  
	\label{fig:RC}   
\end{figure}

RC model is a typical grey-box model and it is widely used in wide range of data-driven modeling applications, for example, describing the thermal dynamics of buildings\cite{ref.RCpaper}. As shown in Fig.\ref{fig:RC}, an building with only one zone in cooling season can be abstracted to three parts consisting of thermal resistances and capacitances. The envelop part is composed of two resistances and one capacitance. More specifically, we only take the wall's thermal dynamic into consideration. As for windows, it can be treated as a resistance. The indoor part represents the property of indoor air, of which the resistance can be ignored. Furthermore, cooling power $Q_c$, windows transmitted beam solar radiation rate $Q_{solar}$ and all heat gains throughout the zone $Q_{in}$ also have great influence on the indoor temperature. The last part is internal mass, which is on behalf of construction/material parameters within the space. It can be viewed as a resistance and a capacitance.

As shown in (\ref{rc_equation}), we can abstract the ordinary differential equations (ODEs) for the thermal characteristics of the RC model based on the  Kirchhoff's law. 
\begin{eqnarray}\label{rc_equation}
C_1\frac{{\rm d}T_{e}}{{\rm d}t}&=&\frac{T_{out}-T_{e}}{R_1}\nonumber\\
C_{in}\frac{{\rm d}T_{in}}{{\rm d}t}&=&\frac{T_{e}-T_{in}}{R_2}-\frac{T_{in}-T{m}}{R_3}\\
& & + Q_{in} + Q_c + Q_{solar}\nonumber\\
C_m\frac{{\rm d}T_{m}}{{\rm d}t}&=&\frac{T_{in}-T_{m}}{R_3}\nonumber
\end{eqnarray}

Given a parameters set , we can solve ODEs (\ref*{rc_equation}) by using the numerical method, for example, fourth-order Runge-Kutta method. Then, we design an objective function in a mean absolute error (MAE) form as shown in (\ref{objective}),
\begin{equation}\label{objective}
f(\bm{x}_{pars}) =\frac{\mid T_{in,sim}(\bm{x}_{pars}) - T_{in,obs}\mid}{N_{obs}}
\end{equation}
where $\bm{x}_{pars}$ is parameters set $[T_{e0}, T_{in0}, T_{m0}, C_1, C_{in},C_m,\\R_1, R_2, R_3]$  to be estimated, and $T_{e0}, T_{in0}, T_{m0}$ are the initial values of $T_e, T_{in}, T_m$. $T_{in,sim}(\bm{x}_{pars})$ represents the numerical results of (\ref{rc_equation}) and $T_{in,obs}$ is the real observation of indoor temperature. $N_{obs}$ is the number of $T_{in,obs}$.

\subsection{B. Results}\label{results}
\begin{figure}[ht]
	\centering  
	\includegraphics[width=0.9\linewidth]{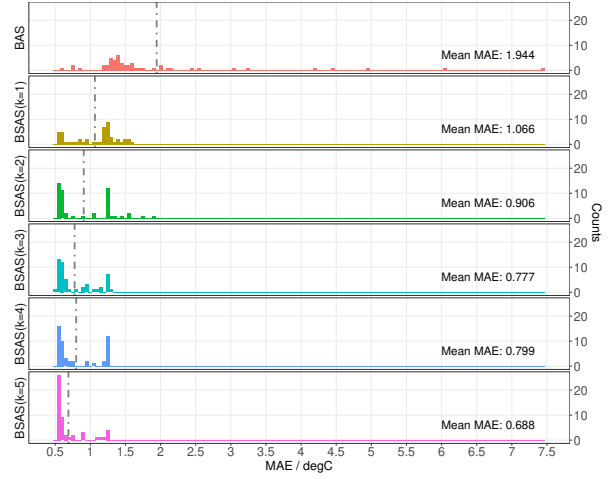} 
	\caption{MAE disturbition of  BAS and BSAS algorithms with different $k$ value in 50 experiments. Mean MAE of BAE's results is absolutely larger than which of BSAS. The figure is generated by ggplot2\cite{ref.ggplot2}. }  
	\label{fig:results}  
\end{figure}

50 experiments are carried out on both of BAS and BSAS algorithm. In addition, we set a range of $k$ from 1 to 5. The results of experiments are shown in Fig.\ref{fig:results}.

The $x$ axis represents convergent results of ibjective function (\ref{objective}) for each experiments. The $y$ axis represents the counts of numerical results within the same intervals. Evidently, BAS performs worst because of the largest mean MAE index. With the $k$ increases, the more compact distribution are obtained and the mean MAE becomes smaller.  BSAS shows better performance than BAS regardless of the $k$ value. It is interesting to note that BSAS with $k = 1$ also performs better than BAS. This phenomenon reveals the efficacy of the \textit{feedback-based step-size update strategy}.

\section{Conclusion}\label{conclusion}

This work presents an extended BAS algorithm to raise BAS's ability and efficency for dealing with high-dimensional problems. Swarm intelligence method and feedback-based step-size update strategy are introduced to modify the original BAS algorithm. System identification for RC model is employed to test the performances of  BSAS in terms of local minimum avoidance. The results reveal the efficacy of proposed algorithm. 

\section{Acknowledgment}\label{acknowledgment}

Gratitude is extended to Dr. Li for his suggestions in algorithm improvement.

\section{Appendix}

A implementation of BAS and BSAS algorithm in form of R package could be found at \url{https://github.com/jywang2016/rBAS}.

\end{document}